\documentclass[10pt,twocolumn,letterpaper]{article}

\usepackage[pagenumbers]{wacv} 

\definecolor{wacvblue}{rgb}{0.21,0.49,0.74}
\usepackage[pagebackref,breaklinks,colorlinks,allcolors=wacvblue]{hyperref}

\def\wacvPaperID{****} 
\def\confName{WACV}
\def\confYear{2026}

\title{SENCA-st: Integrating Spatial Transcriptomics and Histopathology with Cross Attention Shared Encoder for Region Identification in Cancer Pathology}

\author{
Shanaka Liyanaarachchi\(^{\dagger}\) \quad  Chathurya Wijethunga\(^{\dagger}\) \quad Shihab Aaqil Ahamed\(^{\dagger}\) \quad Akthas Absar\(^{\dagger}\) \\
Ranga Rodrigo\(^{\dagger}\) \\
\(^{\dagger}\)Dept. of Electronic and Telecommunication Engineering, University of Moratuwa, Sri Lanka\\
{\tt\small dsglshanaka@gmail.com }
}

\begin{document}
\maketitle
\begin{abstract}
Spatial transcriptomics is an emerging field that enables the identification of functional regions based on the spatial distribution of gene expression. Integrating this functional information present in transcriptomic data with structural data from histopathology images is an active research area with applications in identifying tumor substructures associated with cancer drug resistance. Current histopathology-spatial-transcriptomic region segmentation methods suffer due to either making spatial transcriptomics prominent by using histopathology features just to assist processing spatial transcriptomics data or using vanilla contrastive learning that make histopathology images prominent due to only promoting common features losing functional information. In both extremes, the model gets either lost in the noise of spatial transcriptomics or overly smoothed, losing essential information. Thus, we propose our novel architecture SENCA-st (Shared Encoder with Neighborhood Cross Attention) that preserves the features of both modalities. More importantly, it emphasizes regions that are structurally similar in histopathology but functionally different on spatial transcriptomics using cross-attention. We demonstrate the superior performance of our model that surpasses state-of-the-art methods in detecting tumor heterogeneity and tumor micro-environment regions, a clinically crucial aspect.
\end{abstract}
    
\section{Introduction}
Understanding structural and functional regions in tissues provides insights of underlying pathophysiology of disease states. While histopathology image analysis helps in understanding structural regions, it lacks representation of functional (spatial) regions. Spatial transcriptomics, on the other hand, enables gene expression visualization in space ~\cite{chen2015spatially}, offering a key advantage over bulk transcriptomics by providing spatially resolved functional information. Identification of functional regions, obscure directly in histopathology images, makes spatial transcriptomics a valuable tool specifically for region segmentation in tumor heterogeneity and tumor micro-environment~\cite{lewis2021spatial}.

The concept of tumor heterogeneity arises when rapidly mutating and dividing cancer cells undergo natural selection~\cite{tammela2020investigating}. This consists of a tumor edge that conceals the tumor from immune attacks and drug therapies and a tumor core with invasive cancer. Cellular receptors and genetic makeup of the tumor edge are much like the non-cancer cells and make the tumor indistinguishable while the invasive core rapidly undergoes cancer metastasis~\cite{yu2022spatial}. The tumor micro-environment refers to the surrounding regions of a tumor, which influence the immune system's ability and the effectiveness of therapeutic drugs in targeting the tumor~\cite{li2022spatial}. The segmentation of these regions, the goal of our work, is crucial in clinical planning and testing experimental therapies.

This region identification is challenging due to the presence of thousands of gene channels and undetermined nature of certain gene expression. Early statistical and machine learning based region segmentation attempts relied solely on spatial transcriptomics (unimodal) ~\cite{cang2021scan,richter2023spatialssl,wang2024graph}. These unimodal architectures turned out to be sub-optimal due to transcriptomics inherently being extremely noisy~\cite{xu2022deepst}. Leveraging structural features of histopathology images substantially reduce the effects of noisy nature. This combined with the increased importance of both structural and functional regions led to successful multi-modal approaches~\cite{xu2022deepst}.

Early multi-modal architectures operate at two extremes functional heavy, or structural heavy biasing toward a single modality without a balanced contribution from both modalities. On the functional heavy extreme, SpaGCN \cite{hu2021spagcn} uses structural morphologies as weights of the graphs while DeepST \cite{xu2022deepst} uses structural morphologies to augment spatial transcriptomic data. Both involve minimal inclusion of structural information, and models get confused with noisy spatial transcriptomics data resulting in lower performance. On the other structural-heavy extreme, ConGcR/ConGaR ~\cite{lin2024contrastive} uses contrastive learning between spatial transcriptomic embeddings and histopathology image patch embeddings which cause structural features to dominate the outcome due to the uncontrolled information flow. This drawback calls for a balanced contribution from the special transcriptomic and histopathology image modalities for region segmentation.

In this paper, we present a shared encoder that would learn a fair joint representation of both spatial transcriptomics and histopathology processed in two different branches but control the information flow using a neighborhood cross attention mechanism. This mechanism emphasizes the nodes that have a different correlation ratio of the features of spatial transcriptomics and histopathology compared to their neighbors. This learns a joint representation not dominated by structural features at the local level. We also introduce a hierarchical learning mechanism in which pooled low-resolution data gets trained using contrastive learning, smoothing noise at the global feature level to reduce inherent noise of spatial transcriptomic data but not directly affecting the critical local features. We demonstrate that our model surpasses the state-of-the-art region segmentation results both qualitatively and quantitatively using publicly available data ~\cite{andersson2021spatial,wu2021single, ji2020multimodal}. {Our code is available at \textcolor{blue}{\href{https://github.com/shanaka-liyanaarachchi/SENCA-st}{https://github.com/shanaka-liyanaarachchi/SENCA-st}}}
 
{\bf Our major contributions are}
\begin{enumerate}
    \item Using a novel neighborhood cross-attention shared encoder between histopathology and spatial transcriptomics data leading for better segmentation.
    \item Hierarchical learning in which feature flow from histopathology to spatial transcriptomics controlled at according to resolution.
\end{enumerate}

\section{Related Works}
Machine learning models related to spatial transcriptomics can be divided into two main categories: generative models~\cite{mejia2024enhancing,wang2024cross,chung2024accurate} that predict spatial transcriptomics and inference models that derive insights from spatial transcriptomics data. While many models belong to the first category, our SENCA-st model belongs to the second category of inference models.

Earlier machine learning models existed only using spatial transcriptomics. Using graph neural networks to process spatial transcriptomics data has been suggested in these papers and these papers usually excels at tasks that are strongly presented by multiple gene channels such as segmenting layers of brain cortex. In currently existing multi-modal architectures, a single modality is predominantly determining the output features. SpaGCN ~\cite{hu2021spagcn} use structural morphologies as weights of the spatial transcriptomics graphs and DeepST~\cite{xu2022deepst} use structural morphologies to augment spatial transcriptomic data. ConGcR ~\cite{lin2024contrastive} changes the approach by using the contrastive learning between two modalities.

Previous work has used graph neural networks maximizing mutual information between local node representations and a global summary and they use contrastive learning between augmented versions of feature vector~\cite{cang2021scan,zong2022const}. As these mechanisms do not effectively mitigate the inherent noisiness of spatial transcriptomics data, Lin \etal ~\cite{lin2024contrastive} have suggested using contrastive learning between spatial transcriptomics and histopathology images. However, this approach ultimately results in excessive smoothing, leading to the loss of valuable functional information at the local level. We address this issue by introducing a hierarchical learning mechanism, where smoothing occurs at the global level, while local-level features are learned using a cross-attention shared encoder.

Self-attention was first introduced by Vaswani \etal ~\cite{vaswani2017attention}, primarily for natural language applications, generating attention weights for language tokens using queries, keys, and values derived from the same language sequence. There is an extended version for graphs~\cite{velickovic2017graph}. Self attention has been previously used in uni-modal spatial transcriptomics-only models~\cite{richter2023spatialssl, wang2024graph}. The concept of cross-attention involves generating queries from one modality while generating keys and values from another modality. Previously, attention has been used to relate between different levels of resolution of histopathology images in generative models~\cite{wang2024cross,zhu2025asign}. To the best of our knowledge, we are the first to suggest neighborhood cross-attention to learn multi-modal representation between spatial transcriptomics and histopathology for multi-modal inference.
\section{Method}
\begin{figure*}[t]
\includegraphics[width=\textwidth]{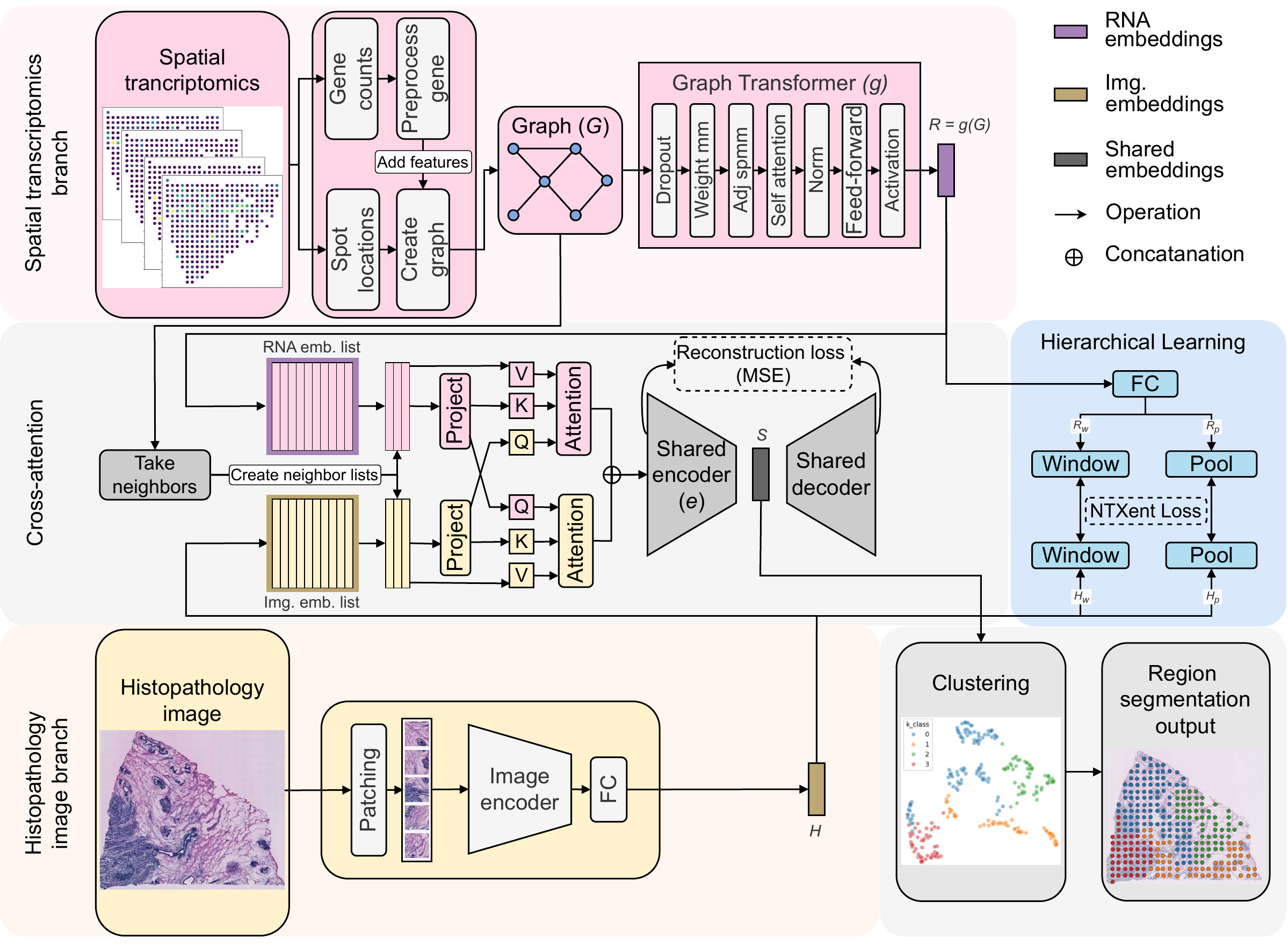}
\caption{SENCA-st system architecture: we leverage both spatial transcriptomics and histopathology data processed through a graph transformer and a ResNet encoder to produce embeddings through neighborhood cross-attention shared encoder which leads to better segmentation.} \label{fig1}
\end{figure*}
In our SENCA-st architecture (Fig.~\ref{fig1})  we have two separate spatial transcriptomics branch and a histopathology branch to generate RNA embeddings(R) and histopathology image embeddings(H). A shared representation learning(S) of these two branches are learn through the cross attention shared encoder that would fairly represent both modalities but also weigh more attention when structurally similar but functionally different regions. This shared embeddings(S) are used to segment out the regions

First, we create a graph $G$ using spatial locations of the spots in which the gene expression has been measured. Expressed genes are then added as the feature vector $V_i$ of each spot (node) $i$~\cite{cang2021scan}. Then, we process the graph through a graph transformer $g$ to generate RNA embeddings ({$R= g(G)$}). Second, in the histopathology image branch, we crop the histopathology image around the spots and pass each image patch through an image encoder to generate image embeddings $H$. These embeddings run through the cross-attention module. 
This operation preserves the features of both modalities and emphasizes regions that are structurally similar in histopathology but functionally different in spatial transcriptomics through cross-attention. Then, we extract the latent space shared embedding $S_i$. This shared embeddings are clustered and the cluster label of the corresponding spot is used to segment the spots.
 We use a weighted combination of Mean Square Error (MSE) between encoder input $E_i$ and decoder output $D_i$ with two NT-Xent contrastive losses proposed in SimCLR~\cite{chen2020simple}, each for windowed correspondence ($H_{w}, R_{w}$) and pooled correspondence ($H_{p}, R_{p}$) of RNA and image embeddings with $\lambda$ to adjust weight.
\begin{equation}
\begin{split}
\mathrm{Loss} = \text{NT-Xent}_{g}(H_{p}, R_{p})+ \text{NT-Xent}_{w}(H_{w}, R_{w}) \\ + \lambda\; \underset{i}{\text{avg}}(\mathrm{MSE}(E_i, D_i)). 
\end{split}
\end{equation} 
The MSE loss intends to learn local features for spots, while temperature-normalized cross entropy $\text{NT-Xent}$ focuses on learning at the global scale. This entire architecture operates in a fully self-supervised and zero-shot fashion. Finally, we benchmark the results using the Adjusted Rand Index (ARI) and compare them to the ground truth label folowing standard becnchmark procedures of literature.

\subsection{Graph Creation for Pre-Processing Transcriptomics:} 
As the first step, we preprocess the spatial transcriptomics data and model it as a graph structure. Following ~\cite{cang2021scan}, we create the spatial graph $G$ for the spots by considering the nearest neighbors based on their physical distance, using the ball-tree algorithm. Then we preprocess the genes by filtering out those with a total count less than 10, normalizing the gene expression, and transforming it to a log scale. Next, we select the top highly variable genes based on their normalized dispersion similar to ConGCR~\cite{lin2024contrastive}. We consider these genes as features, and we add the feature vector {$V_i$} of each spot $i$ to the graph.

\subsection{Transcriptomic Branch---Graph Transformer:} 
Following the pre-processing step in the spatial transcriptomic branch, we use a graph transformer $g$ to process the graph that models the spatial transcriptomic data to learn a lower-dimensional representation of nodes. The graph transformer begins with a dropout layer, which serves more than just preventing overfitting; it also acts as a masking mechanism for a noisy auto-encoder. Next, we perform a matrix multiplication of learnable weights, followed by another spatial matrix multiplication that also considers the adjacency matrix. This process is similar to the message-passing operation in a graph, where information is exchanged between adjacent nodes learning local features. Then the graph passes the output through a transformer block with self-attention, using the same spot embeddings as queries, keys, and values. This allows the model to focus on important nodes in the same way that the original language transformer pays attention to relationships between word embeddings.

Then, we pass the self attention output through layer normalization and a Feed-Forward (FF) network which plays a more significant role in our model than in the original language model. By having a FF network that runs on every node, it also assigns a weight to each gene, in addition to the weights given to nodes by the self-attention layer. Then, a residual connection bypasses the transformer block similar to the original language model~\cite{vaswani2017attention}. Next, we pass the embeddings through an activation layer to produce the final output. The graph transformer ensures that it learns a lower-dimensional representation of nodes while considering their spatial positions, and the attention mechanism helps assign more weight to important nodes.

\subsection{Histopathology Branch - Image Encoder:}
We split the histopathology image into patches by considering a given patch radius (up to third degree neighbor - three times the distance between two spots) around the physical locations of each spot and convert to tensors. Afterwards, we pass these tensors through a ResNet~\cite{he2016deep} image encoder followed by a Fully Connected (FC) layer to generate patch embeddings. We load the ResNet with pre-trained ImageNet~\cite{deng2009imagenet} weights, to provide a basic understanding of visual features, although not specific to histopathology. These encoded path embeddings carry information about the structural features of the tissue spot to the shared encoder $e$.

\subsection{Cross-Attention Shared Encoder:}
We pass the RNA embedding ({$R_i$}) and the image embedding ({$H_i$}) of each spot, along with the RNA embeddings ($R_n$) and image embeddings ($H_n$) of their neighbors (according to the adjacency matrix) through separate projection layers to generate $H_{n,p}$ and $R_{n,p}$. 
The two attention layers follow this. 
For the first attention layer, we generate keys, values, and queries using projected RNA embeddings, original RNA embeddings, and projected image embeddings, respectively. Similarly, for the second attention layer, we generate keys, values, and queries using projected image embeddings, original image embeddings, and projected RNA embeddings, respectively. 
Then, we generate the cross-attention outputs, which weighs embeddings of spots compared to their neighbors. This results in cross-attention between the transcriptomic and histopathology image modalities.  Finally, we extract and concatenate the two vectors for the considered spot. This concatenated vector goes through a dimensionally reducing encoder and then through a dimensionally growing decoder. 
The encoder output corresponding to each spot, $S_i$ leads to the final segmentation.
\begin{equation}
    {S_i=e({A^1_{i}}+A^2_{i})}
\end{equation}
\begin{equation}
    {{A^1_{i} = \mathrm{softmax}\left({(H_{n,p}W^{q1})(R_{n,p}W^{k1})^T}/{\sqrt{E}}\right)R_nW^{v1}}}
\end{equation}
\begin{equation}
    {{A^2_{i} = \mathrm{softmax}\left({(R_{n,p}W^{q2})(H_{n,p}W^{k2})^T}/{\sqrt{E}}\right)H_nW^{v2}}}
\end{equation}
(\(W^{q.}\), \(W^{k.}\),\(W^{v.}\): weights of queries, keys, values respectively and $E$: dimension of the keys vector.)
We use the shared decoder output to train the model while extracting the latent embedding from the encoder as the shared embedding for clustering.

\subsection{Hierarchical Learning:}
We leverage hierarchical learning to effectively learn both local (high resolution) and global (low resolution) features. We use $\text{NT-Xent}$-loss-based contrastive learning at a higher level (low resolution) through windowing and pooling, while the cross-attention shared encoder learns local features (high resolution). The shared encoder slides through every spot and learns a local shared embedding, which we use in clustering. Running contrastive learning at a higher level helps smooth out the noisy nature of spatial transcriptomics controllably. However, it does not directly affect the locally learned shared embedding, as it is done at the global scale, unlike in vanilla contrastive models~\cite{lin2024contrastive}. Finally, we cluster the extracted shared embeddings unsupervised using agglomerative clustering. This approach, which does not use any labels, is  important in looking for unknown regions.
\section{Experiments and Results}
\subsection{Region Identification in Breast Cancer}
\subsubsection{Standardized Testing with HER2ST Dataset}
\begin{figure*}
\includegraphics[width=0.95\textwidth]{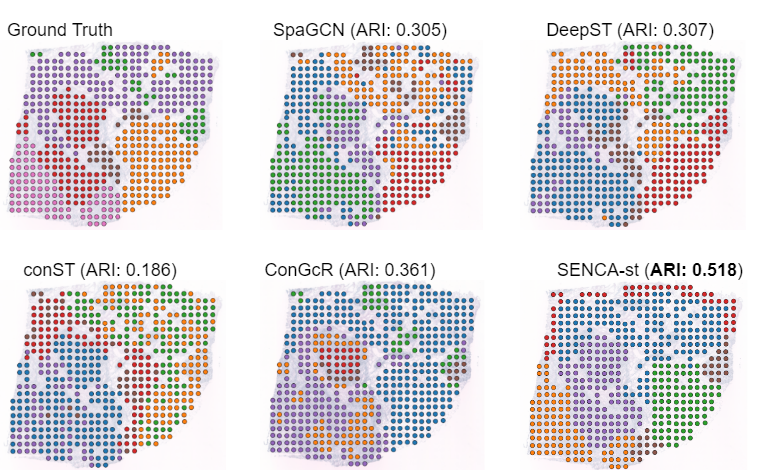}
\caption{Qualitative comparison for H1 sample in HER2ST: SENCA-st (our) clustering closely matches with the ground truth. For this particular sample ARI is high at 0.518.} \label{fig2}
\end{figure*}
\begin{table}
\begin{center}
\caption{HER2ST Full Dataset Results: We report the mean and median ARI results (over the eight annotated samples). Our results significantly surpasses the existing results.}\label{tab2}
\begin{tabular}{@{}lcr@{}}
\toprule
Model &  ARI (Mean) & ARI (Median) \\
\midrule
BayesSpace~\cite{zhao2021spatial}  & { 0.100 } & {0.071}\\
SpaGCN ~\cite{hu2021spagcn}  & { 0.195} & {0.230}\\
DeepST ~\cite{xu2022deepst}& { 0.237} & {0.257}\\
conST ~\cite{zong2022const} & { 0.149} & {0.111}\\
ConGcR ~\cite{lin2024contrastive} & { 0.268} & {0.258}\\
ConGaR ~\cite{lin2024contrastive} & { 0.187} & {0.184}\\
\bfseries SENCA-st (Ours) & {\bfseries 0.304} & {\bfseries 0.320}\\
\bottomrule
\end{tabular}
\end{center}
\end{table}

We used publicly available HER2ST dataset\footnote{Dataset Downloadable from the official github repository of the study.} from a study by Andersson \etal ~\cite{andersson2021spatial} which comprises sections of Human Epidermal Growth Factor Receptor (HER2) positive breast cancer patients. HER2 is a suitable cancer type for targeted therapy~\cite{wang2019targeted,mendes2015benefit}) and susceptible to show resistance to targeted therapies ~\cite{vernieri2019resistance}, where tumor heterogeneity is an important parameter. The dataset contains data for 36 breast cancer sections of 8 patients comprising 
HE-stained images and spatial transcriptomic for each section. There are 8 samples, one per patient, with pathologist's annotations and scientific explanations, which we use for our experiments following Lin \etal \cite{lin2024contrastive}. 

We evaluated all 8 annotated samples independently and calculated the arithmetic mean of the sample ARI (ranges from $-1$ no agreement to $1$ perfect agreement) for the dataset following prior work. ARI measures the similarity between our self-supervised results and ground-truth labels. 
\begin{equation}
\text{ARI} = \frac{\sum_{ij} \binom{n_{ij}}{2} - \left[ \sum_i \binom{a_i}{2} \sum_j \binom{b_j}{2} \middle/ \binom{n}{2} \right]}{\frac{1}{2} \left[ \sum_i \binom{a_i}{2} + \sum_j \binom{b_j}{2} \right] - \left[ \sum_i \binom{a_i}{2} \sum_j \binom{b_j}{2} \middle/ \binom{n}{2} \right]}
\end{equation}
For each sample, we train the model for 10 epochs with a learning rate of {$5 \times 10^{-4}$} using the Adam optimizer \textbf{without prior training on any other sample}. As model parameters, we used 500 as the number of highly variable genes, 128 as the embedding dimension for RNA and image embeddings, 256 as the hidden dimensions of the graph transformer, 4 as the number of neighbors,$\lambda=40$ and  0.2 as the dropout rate. We train our model on a P100 GPU with 16GB of memory.

According to the benchmark results of the previous models from Lin \etal ~\cite{lin2024contrastive}, \textbf{our model outperforms other methods with a substantial margin} (Table~\ref{tab2}). The performance increase in our model is statistically significant ($p<0.05$) with a $p$ value of 0.0055 for the t-test.

Although the quantitative results offer a fundamental understanding of our model's superior performance, its true value lies in the qualitative analysis of real scenarios. In Fig.~\ref{fig2}, the pink cluster in ground truth represents an invasive cancer at the core and the red cluster represents cancer in situ exposing edge. 
\textbf{Ours is the only model that could correctly identify those two regions that are critical in tumor heterogeneity}.\\

\subsubsection{Extended Experiments.}

We conducted extended experiments for other breast cancer types as well since the standard dataset only contained HER2+ breast cancer samples. Those samples \footnote{Dataset Downloadable from the official zenodo repository of the study.} with more than 7 regions were taken from a study by Wu \etal ~\cite{wu2021single}. Those samples used a more information rich technology compared to previous experiment thus had to change embedding dimensions from 256,128 to 512,256 and simultaneously increased input highly variable genes from 500 to 1000. For those samples also each sample, we train the model for 40 epochs with a learning rate of {$5 \times 10^{-4}$} using the Adam optimizer \textbf{without prior training on any other sample}. We report the results in table~\ref{tab3}.

\begin{table}
\begin{center}
\caption{We conducted several extended experiments on other breast cancer types that were not covered in HER2ST dataset. }\label{tab3}
\begin{tabular}{@{}llr@{}}
\toprule
Sample Name & (Type) &  ARI \\
\midrule
CID44971 & (TNBC)  & { 0.326 }\\
CID4535 & (ER+) & { 0.272}\\
1160920F & (TNBC) & { 0.300}\\

\bottomrule
\end{tabular}
\end{center}
\end{table}

\subsection{Region Identification in Squamous Cell Carcinoma}
\begin{table}
\begin{center}
\caption{Marker genes of patient 2 sample of Squamous Cell Carcinoma detected  with Wilcoxon test performed to test the gene is strongly correlated to the particular cluster compared to the rest}\label{tab4}
\begin{tabular}{@{}llr@{}}
\toprule
Gene Name & Cluster &  P-value \\
\midrule
KRT2 & 0  & { $4.771447\times10^{-11}$}\\
FLG & 0 & { $2.019676\times 10^{-07}$}\\
KLK7 & 0 & {$1.831156\times 10^{-03}$}\\
DSC1 & 0 & {$8.774093\times 10^{-03}$}\\
MMP9 & 1 & {$1.095943\times 10^{-10}$}\\
MMP1 & 1 & {$1.095943\times 10^{-10}$}\\
CCL21 & 1 & {$2.575223\times 10^{-08}$}\\
MMP3 & 1 & {$1.880689\times 10^{-07}$}\\
DCD & 2 & {$5.048839\times 10^{-38}$}\\
IGFBP5 & 2 & {$5.310009\times 10^{-31}$}\\
DCN & 2 & {$1.315863\times 10^{-18}$}\\
KRT2 & 3 & {$5.668670\times 10^{-14}$}\\
COL1A1 & 3 & {$1.488934\times 10^{-08}$}\\
COMP & 3 & {$1.488934\times 10^{-08}$}\\
LOR & 3 & {$1.542446\times 10^{-07}$}\\
SPRR1B & 4 & {$2.752813\times 10^{-35}$}\\
S100A7 & 4 & {$1.199169\times 10^{-31}$}\\
SPRR1A & 4 & {$1.281949\times 10^{-30}$}\\
SBSN & 4 & {$1.844224\times 10^{-29}$}\\
DSC2 & 4 & {$4.377951\times 10^{-27}$}\\
IGFBP4 & 5 & {$4.068289\times 10^{-18}$}\\
CCDC80 & 5 & {$8.295423\times 10^{-17}$}\\
DIO2 & 5 & {$8.428684\times 10^{-11}$}\\
FAAP20 & 6 & {$6.318033\times 10^{-05}$}\\
CASP14 & 6 & {$3.996734\times 10^{-03}$}\\

\bottomrule
\end{tabular}
\end{center}
\end{table}
\begin{figure*}
\includegraphics[width=0.98\textwidth]{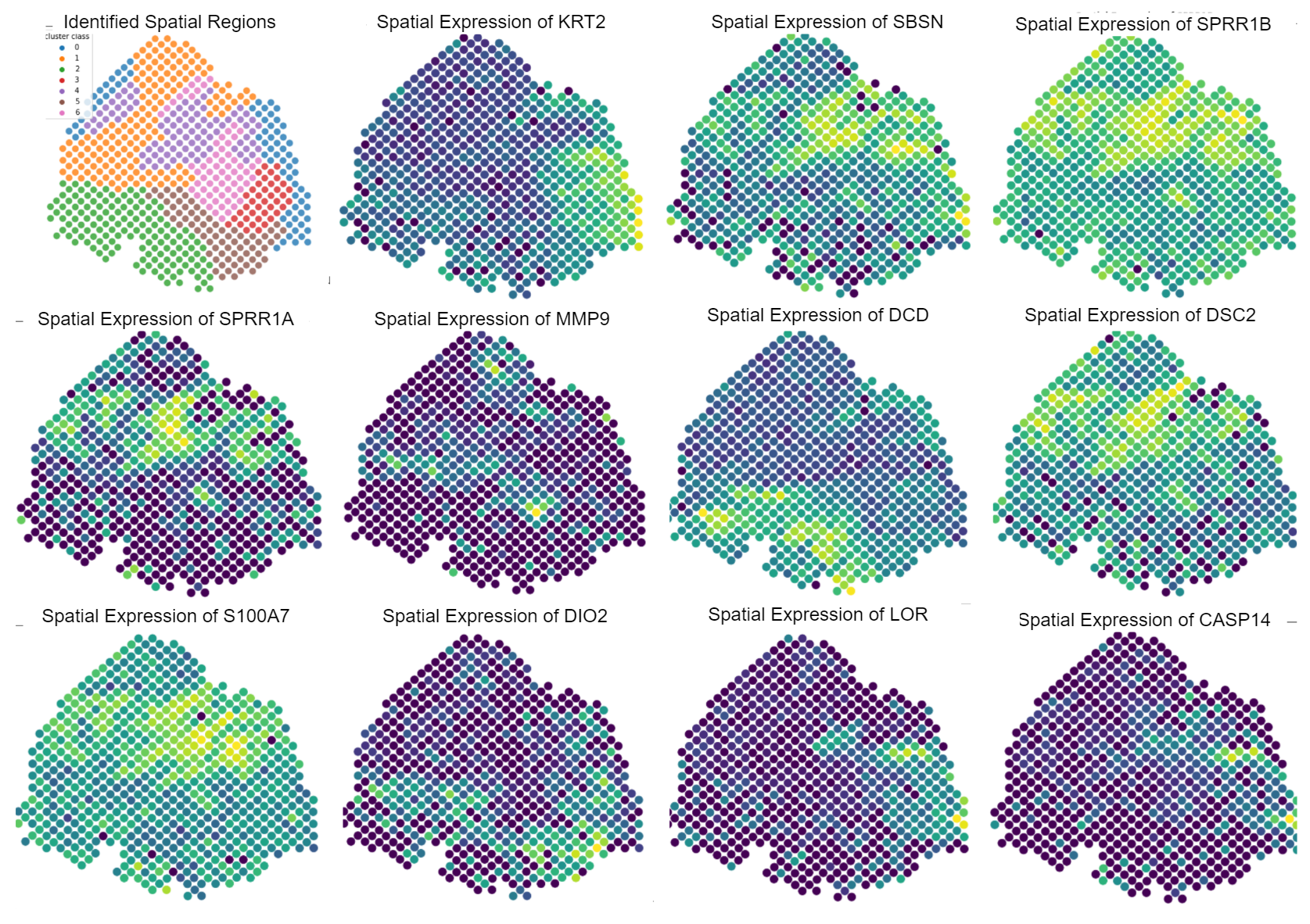}
\caption{Spatial Clusters of sample p2 of Squamous Cell Carcinoma with marker genes.} \label{fig3}
\end{figure*}
\begin{figure*}
\begin{center}
\includegraphics[width=0.78\textwidth]{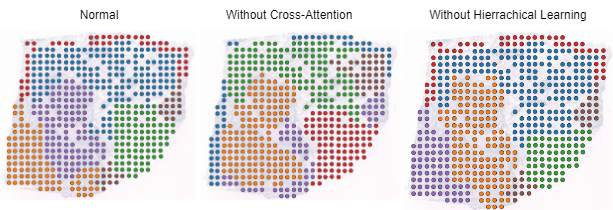}
\caption{Ablation Study - Qualitative effect visualization on H1 sample of HER2ST dataset. We investigated effect of components of the system by removing them and conducting ablation studies.} \label{fig4}
\end{center}
\end{figure*}
We devised our SENCA-st model to identify spatial regions of Squamous Cell Carcinoma samples\footnote{Dataset Downloadable from the official NCBI GEO of the study.} from a study by Ji et. al.~\cite{ji2020multimodal}.
For this experiment also we set model parameters similar to the previous experiment and trained the model for 40 epochs with a learning rate of {$5 \times 10^{-4}$} using the Adam optimizer \textbf{without prior training on any other sample}. We used statistical testing using the Wilcoxon test to find genes correlated with the clusters, and the clusters were extremely correlated with known marker genes. This also means \textbf{that our model could be used to identify biomarkers of unknown pathologies since we did not use any supervision or previous knowledge on the fact that those gene channels are biomarkers when training the model}.

In the experiment with the patient 2 sample correlation with identified marker genes with Wilcoxon test is reported in `Table ~\ref{tab4} and visualization of identified marker genes along with identified regions are presented figure ~\ref{fig3}. Cluster 0 (blue) is statistically significantly($p<0.05$) correlated with KRT2, KRTDAP that are responsible genes for differentiation of  keratinocytes ~\cite{bloor2003expression}. Cluster 0 also had cancer biomarkers such as KLK7 ~\cite{kind2024klk7}, DSC1  ~\cite{myklebust2012expression} statistically significantly correlated to it (Table~\ref{tab4}). Cluster 4 (purple) had cancer biomarkers such as SPRR1B ~\cite{michifuri2013small}, SPRR1A (suggesting tumor is under immune attack with CAF (Cancer-Associated Fibroblast) ~\cite{li2022survival}(Table~\ref{tab4}). Overall both cluster 0 and 4 had SBSN ~\cite{zhou2022sbsn} cancer biomarker (Fig~\ref{fig3}) but statistically more significantly in cluster 4 (Table~\ref{tab4}). Overall cluster 0 and 4 are two distinctive genetically heterogeneous cancer regions.

Cluster 1 (orange) had MMP9, MMP3, MMP1 which are of matrix metalloproteinase (MMP) family that are involved immune activation in tumor micro-environment ~\cite{kessenbrock2010matrix} especially in immune cell recruitment such as Tumor-Associated Macrophages. Cluster 2 (green) has a strong correlation (Table~\ref{tab4}) with DCD ~\cite{schittek2001dermcidin} which a gene prominently expressed in sweat glands and also some variants suggests cachexia (cancer related muscle degradation) \cite{stewart2008dermcidin}. Cluster 3(red) has a strong correlation with LOR as well as KRT2 and Cluster 5(brown) has strong correlation with DIO2~\cite{nappi2025thyroid}. 

This experiment signifies that our model is capable in understanding extremely complicated underlying disease pathology phenomena without getting caught to the inherent noise of spatial transcriptomics. We have covered the full scope of experimentation with achieving the state of the art results with a strong margin in test cases where standard performance indices with annotations are available and statistically showcasing the ability for a self supervised model to understand structural and functional regions statistically matching the existing literature.

\subsection{Ablation Studies}

\begin{table}
\begin{center}
\caption{Ablation Studies - Quantitative Results. We conducted ablation experiments isolating key components of the system by removing them and benchmarking with the HER2ST dataset to study their effect.}\label{tab5}
\begin{tabular}{@{}lr@{}}
\toprule
Ablation Experiment &  ARI (Mean) \\
\midrule
Normal &  { 0.3039 }\\
Without Cross-Attention Weights &  { 0.2218}\\
Without Hierarchical Learning &  { 0.2495}\\
\bottomrule
\end{tabular}
\end{center}
\end{table}
We conducted ablation studies in-order to isolate contribution of each part of the proposed system. Without the cross attention weights ARI droped to 0.2218 and without Hierarchical Learning ARI dropped to 0.2495 from the original 0.3039 signifying the importance of these components as reported in table~\ref{tab5}. Qualitatively without cross attention model was not able to identify structurally similar functionally different regions effectively, when the hierarchical learning was removed cross attention identified the regions but their borders were not close to ground truth like in the original with both the components as seen in the figure~\ref{fig4}.

\subsection{Model Parameters}
\begin{table}
\begin{center}
\caption{Effect of number of neighbors. Even though a small change from actual closest neighbors does not deviate results, large changes make the model confused. This indicate the importance of neighbors}
\label{tab6}
\begin{tabular}{@{}lrrr@{}}
\toprule
 n & n=4 & n=8 & n=16 \\ 
\midrule
ARI(Mean) &  0.304 & 0.290 & 0.240 \\  
\bottomrule
\end{tabular}
\end{center}
\end{table}
We studied the effect of number of neighbors since the spatial transcriptomic grid pattern is uniform square as in Fig~\ref{fig2} number of closest neighbors is 4 which is the recommended number. However when it is increased to 8 including diagonal neighbors other than the orthogonal neighbors results was not deviated much. But when number was increased to 16 which goes beyond immediate neighborhood, results started deviating. We report the results in Table~\ref{tab6}

\subsection{Clustering}
\begin{figure}
\includegraphics[width=0.465\textwidth]{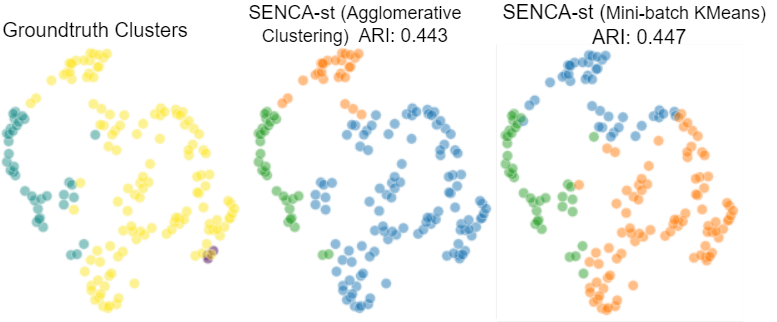}
\caption{Clustering of shared embeddings generated by SENCA-st of C1 sample of HER2ST Dataset} \label{fig5}
\end{figure}

There are several limitations in clustering that could be addressed in future work. First being Optimal number of clusters, even-though we benchmark for annotations for currently known regions there might be regions yet to be discovered and in the Figure~\ref{fig5} we could see some more clusters than the ground truth of three in the UMAP(Uniform Manifold Approximation and Projection). Other limitation is being class size bias in which the small purple cluster not being identified by the algorithm even though embeddings of these spots have been aggregated together. We keep the agglomerative clustering as it produces a dendogram which could be used identify more substructures.

\section{Conclusion}

In this paper, we propose SENCA-st, a neighborhood cross-attention-based shared encoder architecture integrating spatial transcriptomics and histopathology image data. Cross-attention assigns weights to structurally similar but functionally different regions and demonstrates how hierarchical learning diffuses structural features at low resolution without affecting important local functional regions. These novelties have contributed to our architecture performing better than existing work quantitatively and qualitatively. 

We test our model quantitatively with standard benchmarking achieving SOTA performances as well as statistically validating the correlation between identified regions with known biomarkers. More importantly our model demonstrated capabilities to identify critical regions that were not possible to detect with previous methods. Overall we present a novel architecture that address critical issues in existing literature and achieving SOTA results both quantitatively and qualitatively. Future work could focus on the clustering limitations discussed above. We hope that our contributions will have a significant impact on the molecular pathology research community.

\textbf{Acknowledgments} - This project was partially supported by Accelerating Higher Education Expansion and Development (AHEAD) Operation funded by the World Bank.
{
    \small
    \bibliographystyle{ieeenat_fullname}
    \bibliography{main}
}

\end{document}